# An Empirical Study of AI Generated Text Detection Tools

Arslan Akram[1, 2]

[1] Department of Computer Science, Faculty of Computer Science and Information Technology, The Superior University, Lahore, 54000, Pakistan

[2] MLC Lab, Maharban House, House # 209, Zafar Colony, Okara, 56300, Pakistan

*Corresponding Author: Arslan Akram. Email: aaadoula11@gmail.com

**Abstract:** Since ChatGPT has emerged as a major AIGC model, providing high-quality responses across a wide range of applications (including software development and maintenance), it has attracted much interest from many individuals. ChatGPT has great promise, but there are serious problems that might arise from its misuse, especially in the realms of education and public safety. Several AIGC detectors are available, and they have all been tested on genuine text. However, more study is needed to see how effective they are for multi-domain ChatGPT material. This study aims to fill this need by creating a multi-domain dataset for testing the state-of-the-art APIs and tools for detecting artificially generated information used by universities and other research institutions. A large dataset consisting of articles, abstracts, stories, news, and product reviews was created for this study. The second step is to use the newly created dataset to put six tools through their paces. Six different artificial intelligence (AI) text identification systems, including "GPTkit," "GPTZero," "Originality," "Sapling," "Writer," and "Zylalab," have accuracy rates between 55.29 and 97.0%. Although all the tools fared well in the evaluations, originality was particularly effective across the board.

## 1. Introduction

Natural language generation (NLG) models developed more recently have significantly improved the control, variety, and quality of text generated by machines. Phishing [1], disinformation [2], fraudulent product reviews [3], academic dishonesty [4], and toxic spam all take advantage of NLG models' ability to generate novel, manipulable, human-like text at breakneck speeds and efficiencies. A major principle of trustworthy AI [5] addresses the possibility for misuse, such that benefits can be maximized while downsides are minimized. Annie Chechitelli, who serves as Turnitin's chief product officer, provided different instances in which the service discovered false positives. Chechitelli announced on May 14 that the program had processed 38.4 million entries; 9.6% of these submissions featured over 20% artificial intelligence writing, and 3.5% reported employing over 80% artificial intelligence writing [6].

The NLG model is quite robust, but there is room for improvement, such as generating grammatically valid but semantically incoherent or counterfactual language. Even worse, this data can potentially influence public opinion [7], [8]. Thus, language models may be contaminated by mass-produced text [9]. This approach to producing manuscripts can provide fresh and significant challenges to the veracity of scientific publishing and research. Recognizing literature produced by AI saves time for reviewers, contributes to maintaining the integrity of the scientific community, and protects the public from receiving erroneous information. To enhance AI models and human cooperation with AI, researchers need to investigate the gaps between the scientific literature generated by AI and the scientific material published by humans. As a result, we focus on a scenario in which an AI writing helper plays an important role in the scientific community, and we evaluate the quality of scientific literature generated by AI compared to scientific language authored by humans.

In recent years, a lot of interest has been directed toward the potential of generative models such as ChatGPT to generate human-like writing, images, and other forms of media. An OpenAI-developed variant of the ubiquitous GPT-3 language model, ChatGPT is designed specifically for generating text suitable for conversation and may be trained to perform tasks including answering questions, translating text, and creating new languages [10]. Even though ChatGPT and other generative models have made great advances in creating human-like language, it still needs to be determined the difference between writing generated by a machine and text written by a human. This is the case even though ChatGPT and other generative models. This is of the utmost importance in applications such as content moderation, where it is important to identify and remove hazardous information and automated spam [11].

Recent work has centered on enhancing pre-trained algorithms' capacity to identify text generated by artificial intelligence. Along with GPT-2, OpenAI also released a detection model consisting of a RoBERTa-based binary classification system that was taught to distinguish between human-written and GPT-2-generated text. Integrating source-domain data with in-domain labeled data is what Black et al. (2021) do to overcome the difficulty of finding GPT-2-generated technical research literature [12]. The challenge and dataset on detecting machine-created scientific publications, DagPap22, were proposed by Kashnitsky et al. [13]. During the COLING 2022 session on Scholarly Document Processing. Algorithms like GPT-3, GPT-neo, and led-large-book-summary are examples of abstract algorithms used. DagPap22's prompt templates necessitate including information on the primary topic and scientific structural function, making it more probable that the tool will collect problematic and easily-discoverable synthetic abstracts [14], [15]. More recently, GPTZero has been proposed to detect ChatGPT-generated text, primarily based on perplexity. Recent studies have revealed two major issues that need addressing. To begin, every study given here had to make do with small data samples. Thus, a larger, more robust data set is required to advance our understanding. Second, researchers have typically used mock data to fine-tune final versions of pre-train models. Text created with various artificial intelligence programs should all be detectable by the same approach.

Recently developed algorithms for detecting AI-generated text can tell the difference between the two. These systems employ state-of-the-art models and algorithms to decipher text created by artificial intelligence. One API that can accurately identify AI-generated content is Check For AI, which analyses text samples. A further tool called Compilatio uses sophisticated algorithms to identify instances of plagiarism, even when they are present in AI-generated content. Similarly, Content at Scale assesses patterns, writing style, and other language properties to spot artificially generated text. Crossplag is an application programming interface (API) that can detect AI-generated text in a file. The DetectGPT artificial intelligence content detector can easily identify GPT model-generated text. We use Go Winston to identify artificially manufactured news content and social media content. Machine learning and linguistic analysis are used by GPT Zero to identify AI-generated text. The GPT-2 Output Detector Demo over at OpenAI makes it simple to test if a text was produced by a GPT-2 model. OpenAI Text Classifier uses an application programming interface to categorize text, including text generated by artificial intelligence. In addition to other anti-plagiarism features, PlagiarismCheck may identify information created by artificial intelligence. Turnitin is another well-known tool that uses AI-generated text detection to prevent plagiarism. The Writeful GPT Detector is a web-based tool that uses pattern recognition to identify artificially produced text. Last but not least, the Writer can spot computer-generated text and check the authenticity and originality of written materials. Academics, educators, content providers, and businesses must deal with the challenges of AI-generated text, but new detection techniques are making it easier.

This research will conduct a comparative analysis of AI-generated text detection tools using self-generated custom dataset. To accomplish this, the researchers will collect datasets using a variety of artificial intelligence (AI) text generators and humans. This study, in contrast to others that have been reported, incorporates a wide range of text formats, sizes, and organizational patterns. The next stage is to test and compare the tools with the proposed tool. The following bullet points present the most significant takeaways from this study's summary findings.

- Collecting dataset using different LLMs on different topics having different sizes and writing styles.
- Investigation of detection tools for AI text detection on collected dataset.
- Comparison of the proposed tool to other cutting-edge tools to demonstrate the interpretability of the best tool among them.

The following is the structure of this article: The results and comments are presented in Section 4, while Section 2 provides a brief literature review. The final chapter summarizes the work and recommends where the authors could go.

## 2. Literature Review

Artificial intelligence-generated text identification has sparked a paradigm change in the ever-evolving fields of both technology and literature. This innovative approach arises from the combination of artificial intelligence and language training, in which computers are given access to reading comprehension strategies developed by humans. Like human editors have done for decades in the physical world, AI-generated text detection is now responsible for determining the authenticity of literary works in the digital world. The ability of algorithms to tell the difference between human-authored material and that generated by themselves is a remarkable achievement of machine learning. As we venture into new waters, there are serious implications for spotting plagiarized work, gauging the quality of content, and safeguarding writers' rights. AI-generated text detection is a guardian for literary originality and reassurance that the spirit of human creativity lives on in future algorithms, connecting the past and future of written expression.

In their detailed analysis of three custom-built LLMs on a dataset they created, ChatGPT Comparison Corpus (HC3), Guo et al. [16] Using F1 scores for each corpus or sentence, we evaluated the performance of the three models and found that the best model had a maximum F1 score of 98.78%. Results indicated superiority over other SOTA strategies. Because the authors of the corpus only used abstract paragraphs from a small subset of the available research literature, the dataset is skewed toward that subset. The presented models may need to perform better on a general-purpose data set. Wang et al. [17] have offered a benchmarked dataset-based comparison of several AI content detection techniques. For evaluation, they have used question-and-answer, code-summarization, and code-generation databases. The study found an average AUC of 0.40 across all selected AI identification techniques, with the datasets used for comparison containing 25k samples for both human and AI-generated content. Due to the lack of diversity in the dataset, the chosen tools performed poorly; a biased dataset cannot demonstrate effective performance. Tools can also be more accurate or have a higher area under the curve (AUC).

Catherine et al. [18] employed the 'GPT-2 Output Detector' to assess the quality of generated abstracts. The study's findings revealed a significant disparity between the abstracts created and the actual abstracts. The AI output detector consistently assigned high 'false' scores to the generated abstracts, with a median score of 99.98% (interquartile range: 12.73%, 99.98%). This suggests a strong likelihood of machine-generated content. The initial abstracts exhibited far lower levels of 'false' ratings, with a median of 0.02% and an interquartile range (IQR) ranging from 0.02% to 0.09%. The AI output detector exhibited a robust discriminatory ability, evidenced by its AUROC (Area Under the Receiver Operating Characteristic) value of 0.94. Utilizing a website and iThenticate software to conduct a plagiarism detection assessment revealed that the generated abstracts obtained higher scores, suggesting a greater linguistic similarity to other sources. Remarkably, human evaluators had difficulties in distinguishing between authentic and generated abstracts. The researchers achieved an accuracy rate of 68% in accurately identifying abstracts generated by ChatGPT. Notably, 14% of the original abstracts were produced by machine-generated methods. The literature critiques have brought attention to the issue of abstracts that are thought to be generated by artificial intelligence (AI).

Using a dataset generated by the users themselves, Debora et al. [19] compared and contrasted multiple AI text detection methods. The research compared 12 publicly available tools, two proprietary and available only to qualified academic institutions and other research groups. The researchers' primary focus has been explaining why and how artificial intelligence (AI) techniques are useful in the academy and the sciences. The results of the comparison were then shown and discussed. Finally, the limitations of AI technologies regarding evaluation criteria were discussed.

To fully evaluate such detectors, the team [20] first trained DIPPER, a paraphrase generation model with 11 billion parameters, to rephrase entire texts in response to contextual information such as user-generated cues. Using scalar controls, DIPPER's paraphrased results can be tailored to vocabulary and sentence structure. Extensive testing proved that DIPPER's paraphrase of AI-generated text could evade watermarking techniques and GPTZero, DetectGPT, and OpenAI's text classifier. The detection accuracy of DetectGPT was decreased from 70.3% to 4.6% while maintaining a false positive rate of 1% when DIPPER was used to paraphrase text generated by three well-known big language models, one of which was GPT3.5-davinci-003. These rephrases were impressive since they didn't alter the original text's meaning. The study developed a straightforward defense mechanism to safeguard AI-generated text identification from paraphrase-based attacks. Language model API providers were required to get semantically identical texts for this defense strategy to work. To find sequences comparable to the candidate text, the algorithm looked through a collection of already generated sequences. A 15-million-generation database derived from a finely tuned T5-XXL model confirmed the efficacy of this defense strategy. The software identified Paraphrased generations in 81% to 97% of test cases, demonstrating its efficacy. Remarkably, only 1% of human-written sequences were incorrectly labeled as AI-generated by the software. The project made its code, models, and data publicly available to pave the way for additional work on detecting and protecting AI-generated text.

OpenAI [21], an AI research company, compared manual and automatic ML-based synthetic text recognition methods. Utilizing models trained on GPT-2 datasets enhances the inherent authenticity of the text created by GPT-2, hence facilitating human evaluators' identification of erroneous datasets. Consequently, the team evaluated a rudimentary logistic regression model, a detection model based on fine-tuning, and a detection model employing zero-shot learning. A logistic regression model was trained using TFIDF, unigram, and bigram features and evaluated using various generating processes and model parameters afterward. The most basic classifiers demonstrated an accuracy rate of 97% or higher. Models need help in identifying shorter outputs. Topological Data Analysis (TDA) was utilized by Kushnareva et al. [22] to count graph components, edges, and cycles. Text recognition machine learning used these features. The characteristics trained a logistic regression classifier on WebText, Amazon Reviews, RealNews, and GROVER [23]. ChatGPT's lack of thorough testing makes this approach's success uncertain.

The online application DetectGPT was used to zero-shot identify and separate AI-generated text from human-generated text in another investigation [24]. Log probabilities from the generative model were employed. The researchers found intentionally generated text in the model's log probability function's negative curvature. The authors thought assessing the log probability of the models under discussion was always possible. This method only works with GPT-2 cues, the scientists say. In another study, Mitrovic et al. trained an ML model to identify ChatGPT queries from human ones [25]. ChatGPT-generated two-line restaurant reviews were recognized by DISTILBERT, a lightweight BERT-trained and Transformer-tuned model. SHAP explained model predictions. Researchers observed that the ML model couldn't recognize ChatGPT messages. The authors introduced AICheatCheck, a web-based AI detection tool that can distinguish if a text was produced by ChatGPT or a human [26]. AICheck analyzes text patterns to detect origin. The writers used Guo et al. [16] and education to make do with limited data. The study must explain AICheatCheck's precision. The topic was recently investigated by Cotton et al. [27]. The benefits and drawbacks of using ChatGPT in the classroom concerning plagiarism are discussed. In another text [28], the authors used statistical distributions to analyze simulated data. Using a GLTR application, they make sure the text you put in is correct by highlighting it in different colors. Questions used on the GLTR exam were written by the general public and based on the publicly accessible GPT-human-generated content for the 21.5B parameter model [10]. The authors also studied human subjects by having students spot instances of fabricated news.

Different AI-generated text classification models have been presented in recent years, with approaches ranging from deep learning and transfer learning to machine learning. Furthermore, software incorporated the most effective models to help end users verify AI-generated writing. Some studies evaluate various AI text detection tools by comparing their performance on extremely limited and skewed datasets. Therefore, it is necessary to have a dataset that includes samples from many domains written in the same language that the models were trained in. It needs to be clarified which of the many proposed tools for AI text identification is the most effective. To find the best tool for each sort of material, whether from a research community or content authors, it is necessary to do a comparative analysis of the top-listed tools.

## 3. Material and Methods

Many different tools for recognizing artificial intelligence-created text were compared in this analysis. The approach outlined here consists of three distinct phases. Examples of human writing are collected from many online sources, and OpenAI frameworks are used to generate examples of AI writing from various prompts (such as articles, abstracts, stories, and comment writing). In the following stage, you will select six applications for your newly formed dataset. Finally, the performance of the tools is provided based on several state-of-the-art measurements, allowing end users to pick the best alternative. Figure 1 depicts the overall structure of the executing process.

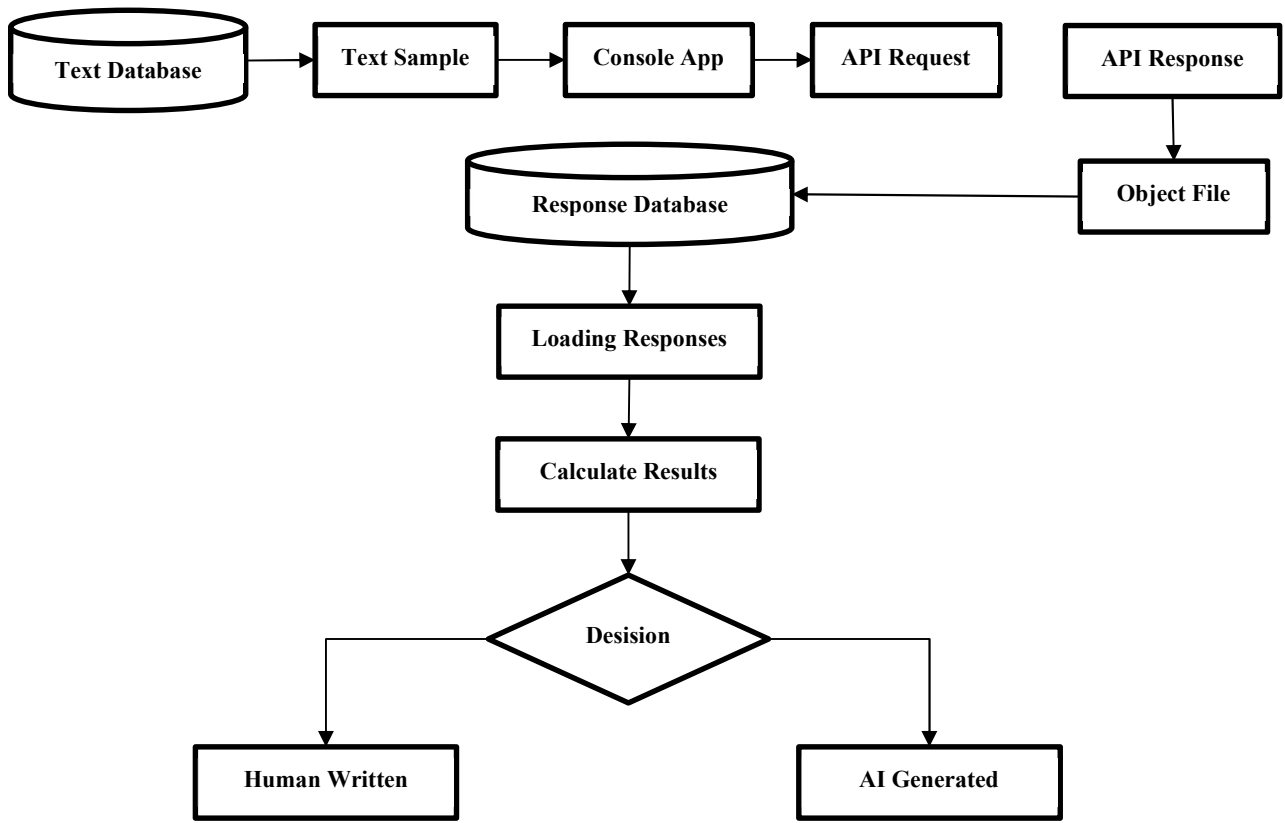

**Figure 1:** Implementation framework for testing tools

*3.1 Datasets*

The primary goal of this study is to amass human-written samples from many sources, including academic databases such as Google Scholar and Research Gate, content producer and blogger databases such as Wikipedia, and other knowledge aggregators. The dataset collected using above mentioned tools is named as AH&AITD (Arslan's Human and AI Text Database) is available at this link (). The samples used for testing are divided into two groups in Table 1: "Human Written" and "AI Generated." The "Human Written" section of the dataset is further broken down into subheadings like "Open Web Text," "Blogs," "Web Text," "Q&A," "News Articles," "Opinion Statements," and "Scientific Research." The number of samples from each source is also included in this group's total of 5,790 samples. The "AI-Generated" group, on the other hand, has a wide range of AI models, each with its own sample size (such as ChatGPT [29], GPT-4 [30],

Paraphrase [31], GPT-2 [32], GPT-3 [33], DaVinci, GPT-3.5, OPT-IML [34], and Flan-T5 [35]). The final number in the table is 11,580, the total number of students in both groups. This table is useful since it shows where the testing dataset came from and how the AI models were evaluated. Table 1 presents details of testing samples used for evaluation of targeted AI generated text detection tools.

**Table 1:** Testing dataset samples (AH&AITD).

| Class | Source | Number of Samples | Total |
|---|---|---|---|
| Human Written | Open Web Text | 2343 | 5790 |
| | Blogs | 196 | |
| | Web Text | 397 | |
| | Q&A | 670 | |
| | News Articles | 430 | |
| | Opinion Statements | 1549 | |
| | Scientific Research | 205 | |
| AI Generated | ChatGPT | 1130 | 5790 |
| | GPT-4 | 744 | |
| | Paraphrase | 1694 | |
| | GPT-2 | 328 | |
| | GPT-3 | 296 | |
| | Davinci | 433 | |
| | GPT-3.5 | 364 | |
| | OPT-IML | 406 | |
| | Flan-T5 | 395 | |
| **Total** | | **11580** | **11580** |

The method of data collecting for human written samples is based on a variety of various approaches. Most human-written samples were manually obtained from human-written research articles, and only abstracts were used. Additionally, open web texts are harvested from various websites, including Wikipedia.

### 3.2 AI Generated Text Detection Tools

Without AI-generated text detection systems, which monitor automated content distribution online, the modern digital ecosystem would collapse. These systems employ state-of-the-art machine learning and natural language processing methods to identify and label data generated by AI models like GPT-3, ChatGPT, and others. They play a vital role in the moderation process by preventing the spread of misinformation, protecting online communities, and identifying and removing fake news. Platform administrators and content moderators can use these tools to spot literature generated by artificial intelligence by seeing patterns, language quirks, and other telltale signals. The importance of AI-created text detection tools for user security, ethical online discourse, and legal online material has remained strong despite the advancements in AI. In this section, AI generated text detection tools are described briefly.

*3.2.1 AI Text Detection API (Zylalab)*

Zyalab's AI Text Detection API [36] makes locating and analyzing text in various content types simple. This API employs cutting-edge artificial intelligence (AI) technology to precisely recognize and extract textual content from various inputs, including photos, documents, and digital media. AI Text Detection API uses cutting-edge OpenAI technology to identify ChatGPT content. Its high accuracy and simple interface let instructors spot plagiarism in student essays and other AI-generated material. Its ease of integration into workflows and use by non-technical users is a major asset. Due to OpenAI's natural language processing powers, the API can detect even mild plagiarism, ensuring the information's uniqueness. It helps teachers grade essays by improving the efficiency of checking student work for originality. In conclusion, the AI Text Detection API simplifies, accurately, and widely applies plagiarism detection and essay grading for content suppliers, educators, and more. Due to its ability to analyze text and provide a detailed report, this tool can be used for plagiarism detection, essay grading, content generation, chatbot building, and machine learning research. There are no application type constraints, merely API request limits. The API makes use of OpenAI technology. It has a simple interface and high accuracy, allowing it to detect plagiarism in AI-generated writing and serve as an essay detector for teachers.

*3.2.2 GPTKIT*

The innovators of GPTKit [37] saw a need for an advanced tool to accurately identify Chat GPT material, so they built one. GPTKit is distinguished from other tools because it utilizes six distinct AI-based content recognition methods, all working together to considerably enhance the precision with which AI-generated content may be discovered. Educators, professionals, students, content writers, employees, and independent contractors worried about the accuracy of AI-generated text will find GPTKit highly adaptable. When users input text for analysis, GPTKit uses these six techniques to assess the content's authenticity and accuracy. Customers can try out GPTKit's features for free by having it return the first 2048 characters of a response to a request. Due to the team's dedication to continuous research, the detector in GPTKit now claims an impressive accuracy rate of over 93% after being trained on a big dataset. You can rest easy knowing that your data will remain private during detection and afterward, as GPTKit only temporarily stores information for processing and promptly deletes it from its servers. GPTKit is a great tool to use if you wish to validate information with artificial intelligence (AI) for authenticity or educational purposes.

*3.2.3 GPTZero*

GPTZero is the industry standard for identifying Large Language Model documents like ChatGPT. It detects AI content at the phrase, paragraph, and document levels, making it adaptable. The GPTZero model was trained on a wide range of human-written and AI-generated text, focusing on English prose. After servicing 2.5 million people and partnering with 100 education, publishing, law, and other institutions, GPTZero is a popular AI detector. Users may easily enter text for analysis using its simple interface, and the system returns detailed detection findings, including sentence-by-sentence highlighting of AI-detected material, for maximum transparency. GPTZero supports numerous AI language models, making it a versatile AI detection tool. ChatGPT, GPT-4, GPT-3, GPT-2, LLaMA, and AI services are included. It was the most accurate and trustworthy AI detector of seven tested by TechCrunch. Customized for student writing and academic prose, GPTZero is ideal for school. Despite its amazing powers, GPTZero [38] admits it has limitations in the ever-changing realm of AI-generated entertainment. Thus, teachers should combine its findings into a more complete assessment that prioritizes student comprehension in safe contexts. To help teachers and students address AI misuse and the significance of human expression and real-world learning, GPTZero emphasizes these topics. In-person evaluations, edited history analysis, and source citations can help teachers combat AI-generated content. Due to its commitment to safe AI adoption, GPTZero may be a reliable partner for educators facing AI issues.

*3.2.4 Sapling*

Sapling AI Content Detector [39] is a cutting-edge program that accurately recognizes and categorizes AI-generated media. This state-of-the-art scanner utilizes state-of-the-art technology to verify the authenticity and integrity of text by checking for the existence of AI-generated material in various contexts. Whether in the courtroom, the publishing industry, or the classroom, Sapling AI Content Detector is a potent solution to the issue of AI-generated literature. Its straightforward interface and comprehensive detection results equip users to make informed judgments about the authenticity of the material. Sapling AI Content Detector's dedication to precision and dependability makes it a valuable resource for companies and individuals serious about preserving the highest possible content quality and originality requirements.

*3.2.5 Originality*

The Originality AI Content Detector [40] was intended to address the growing challenge of identifying AI-generated text. This cutting-edge artificial intelligence influence detector can tell if human-written content has been altered. It examines every word, every sentence, every paragraph, and every document. Human-made and computer-generated texts may be distinguished with confidence thanks to the rich variety of training data. Educators, publishers, academics, and content producers will find this tool invaluable for guarding the authenticity and integrity of their own work. The Originality AI Content Detector highlights potential instances of AI-generated literature to increase awareness and promote the responsible use of AI technologies in writing. The era of AI-driven content creation gives users the knowledge to make purposeful decisions that preserve the quality and originality of their writing.

*3.2.6 Writer*

The Writer AI Content Detector [41] is cutting-edge software for spotting content created by artificial intelligence. This program utilizes cutting-edge technologies to look for signs of artificial intelligence in the text at the phrase, paragraph, and overall document levels. Since he was taught using a large dataset, including human-authored and AI-generated content, the Writer is very good at telling them apart. This guide is a must-read for every instructor, publisher, or content provider serious about their craft. By alerting users to the presence of AI content and offering details about it, the Writer arms them with the information they need to protect the authenticity of their works. The author is an honest champion of originality, advocating for responsible and ethical content generation in an era when AI is increasingly involved in the creative process. Developers can get the Writer AI Content Detector SDK by running "pip install writer." The use of API credentials for writer authentication is critical [42]. These API keys can be found on your account's dashboard. Simply replace the sample API keys in the code snippets with your own or sign in for individualized code snippets. Users without access to their secret API keys on the control panel. To become a Writer account's development team member, you should contact the account's owner. Developers can access the Writer SDK and AI Content Detector once signed in. The SDK includes document and user management tools, content identification, billing information retrieval, content production, model customization, file management, snippet handling, access to a style guide, terminology management, user listing, and management. With this full suite of resources, customers can confidently include AI-driven content recognition into their projects and apps without compromising safety or precision.

*3.3 Experimental Setup*

Six distinct content identification approaches developed using artificial intelligence were evaluated in depth for this study. Each tool has an API that can be used with various languages and frameworks. To take advantage of these features, subscriptions have been obtained for each API, and the software has been put through its pace with Python scripts. The results were produced using the testing dataset discussed above. All experiments have been run on a sixth-generation Dell I7 system with 24 GB of RAM and 256 SSD ROM using Python 3.11 on MS Code with Jupyter Notebook Integration.

*3.3 Evaluation Policy*

To ensure the robustness, dependability, and usefulness of a company's machine-learning models, the company should develop and adhere to an evaluation policy. This policy spells evaluation, validation, and application of models in detail. As a first step, it converges on a standardized approach to evaluation, allowing for fair and uniform assessment of model performance across projects. Comparing projects, identifying best practices, and maximizing model development are all made easier with the introduction of uniform standards. Second, a policy for assessing model performance guarantees that they hit targets for measures like accuracy, precision, and recall. As a result, only high-quality, reliable models with strong performance are deployed. Reduced implementation risks are achieved through the policy's assistance in identifying model inadequacies, biases, and inaccuracies. An assessment policy fosters accountability and trustworthiness in data science by requiring uniformity and transparency in model construction.

Accuracy is important in machine learning and statistics because it measures model prediction. Accuracy is a percentage of accurately predicted cases to the dataset's total occurrences. The term "accuracy" could mean:

$$Accuracy = \frac{Number\ of\ correct\ predictions}{Total\ Number\ of\ predictions} \times 100$$

In this formula, the "Total Number of Predictions" represents the size of the dataset, while the "Number of Correct Predictions" is the number of predictions made by the model that corresponds to the actual values. A quick and dirty metric to gauge a model's efficacy is accuracy, but when one class greatly outnumbers the other in unbalanced datasets, this may produce misleading results.

Precision is the degree to which a model correctly predicts the outcome. In the areas of statistics and machine learning, it is a common metric. The number of correct positive forecasts equals the ratio of true positive predictions to all positive predictions. The accuracy equation can be described as follows:

$$Precision = \frac{True\ Positives}{True\ Positives + False\ Positives}$$

The avoidance of false positives and negatives in practical use is what precision quantifies. A high accuracy score indicates that when the model predicts a positive outcome, it is more likely to be true, which is especially important in applications where false positives could have major consequences, such as medical diagnosis or fraud detection.

Recall (true positive rate or sensitivity) is an important performance metric in machine learning and classification applications. It measures a model's ability to discover and label every instance of interest in a given dataset. To recall information, follow this formula:

$$Recall = \frac{True\ Positives}{True\ Positives + False\ Negatives}$$

In this formula, TP represents the total number of true positives, whereas FN represents the total number of false negatives. Medical diagnosis and fraud detection are two examples of areas where missing a positive instance can have serious effects; applications with a high recall, which indicates the model effectively catches a large proportion of the true positive cases, could profit greatly from such a model.

The F1 score is a popular metric in machine learning that combines precision and recall into a single value, offering a fairer evaluation of a model's efficacy, especially when working with unbalanced datasets. The formula for its determination is as follows:

$$F1\ Score = 2 \times \frac{(Precision \times Recall)}{(Precision + Recall)}$$

Precision is the proportion of correct predictions relative to the total number of correct predictions made by the model, whereas recall measures the same proportion relative to the number of genuine positive cases in the dataset. The F1 score excels when a compromise between reducing false positives and false negatives is required, such as medical diagnosis, information retrieval, and anomaly detection. By factoring in precision and recall, F1 is a well-rounded measure of a classification model's efficacy.

A machine learning classification model's accuracy can be evaluated using the ROC curve and the Confusion Matrix. The ROC curve compares the True Positive Rate (Sensitivity) to the False Positive Rate (1-Specificity) at different cutoffs to understand a model's discriminatory ability. The Confusion Matrix provides a more detailed assessment of model accuracy, precision, recall, and F1-score, which meticulously tabulates model predictions into True Positives, True Negatives, False Positives, and False Negatives. Data scientists and analysts can use these tools to learn everything they need to know about model performance, threshold selection, and striking a balance between sensitivity and specificity in classification jobs.

## 4. Results and Analysis

The results and discussion surrounding these tools reveal intriguing insights into the usefulness and feasibility of six AI detection approaches for differentiating AI-generated text from human-authored content. Detection technologies, including GPTZero, Sapling, Writer, AI Text Detection API (Zyalab), Originality AI Content Detector, and GPTKIT, were ranked based on several factors, including accuracy, precision, recall, and f1-score. Table 2 compares different AI text detection approaches that can be used to tell the difference between AI-written and -generated text.

**Table 1:** Comparative results of AI text detection tools on AH&AITD.

| Tools | Classes | Precision | Recall | F1 Score |
|---|---|---|---|---|
| GPTKIT | AI Generated | 90 | 12 | 21 |
| | Human Written | 53 | 99 | 69 |
| GPTZERO | AI Generated | 65 | 60 | 62 |
| | Human Written | 63 | 68 | 65 |
| Originality | AI Generated | **98** | **96** | **97** |
| | Human Written | **96** | **98** | **97** |
| Sapling | AI Generated | 86 | 40 | 54 |
| | Human Written | 61 | 94 | 74 |
| Writer | AI Generated | 79 | 52 | 62 |
| | Human Written | 64 | 87 | 74 |
| Zylalab | AI Generated | 84 | 45 | 59 |
| | Human Written | 62 | 91 | 74 |

First, GPTKIT impresses with its high F1 Score (21) because of its high precision (90) in detecting human-written text but shockingly low recall (12). This suggests that GPTKIT is overly conservative, giving rise to several false negatives. On the other hand, its recall (99) and F1 Score (69) are excellent when recognizing language created by humans. GPTZero's performance is more uniformly excellent across the board. Its recall (60%) and F1 Score (62) more than make up for its lower precision (65%) on AI-generated text. An F1 Score of 65 for human written text strikes a reasonable balance between accuracy (63) and recall (68).

When distinguishing AI-generated content from machine-written content, Originality shines. Its F1 Score of 97 reflects its remarkable precision (98), recall (96), and overall effectiveness. It also excels at text created by humans, with an F1 Score of 97, recall of 98%, and precision of 96%. The high precision (86) and recall (40) on AI-generated text give Sapling an F1 Score of 54. Despite a high recall (94) and poor precision (61) in identifying human-written text, the F1 Score 74 leaves room for improvement. Writer is unbiased in assessing the relative merits of AI-generated and human-written content. It has an average F1 Score of 62 because it has an average level of precision while analyzing AI-generated text (79) and recall (52). The F1 Score for this piece of human-written text is 74, meaning it has an excellent balance of precision (64) and recall (87). Regarding recognizing AI-generated content, Zylalab has a good 84 precision, 45 recall, and 59 F1 Score. Recognizing synthetic language is where it shines, with an F1 Score of 74, a recall of 91, and a precision of 62.

As a result of its superior performance in terms of precision, recall, and F1 Score across both classes, we have concluded that Originality is the most reliable alternative for AI text identification. Additionally, GPTZERO displays all-around performance, making it a practical option. However, Sapling shows skills in identifying AI-generated text whereas GPTKIT demonstrates remarkable precision but needs better recall.

Writers find a comfortable medium ground but need to differentiate themselves. Zylalab performs about as well as the best of the rest, but it has room to grow. Before selecting a tool, it is crucial to consider the needs and priorities of the job.

Figure 2 provides a visual representation of a comparison between the accuracy of six different AI text identification systems, including "GPTkit," "GPTZero," "Originality," "Sapling," "Writer," and "Zylalab." Data visualization demonstrates that "GPTkit" has a 55.29 percent accuracy rate, "GPTZero" has a 63.7 percent accuracy rate, "Originality" has a spectacular 97.0 percent accuracy rate, "Sapling" has a 66.6 percent accuracy rate, "Writer" has a 69.05 percent accuracy rate, and "Zylalab" has a 68.23 percent accuracy rate. These accuracy ratings demonstrate how well the tools distinguish between natural and computer-generated text. When contrasting the two forms of writing, "Originality" achieves the highest degree of accuracy. Compared to the other two, "GPTkit" has the lowest detection accuracy and thus the most room for improvement. This visual representation of the performance of various AI text detection tools will be an important resource for users looking for the most precise tool for their needs.

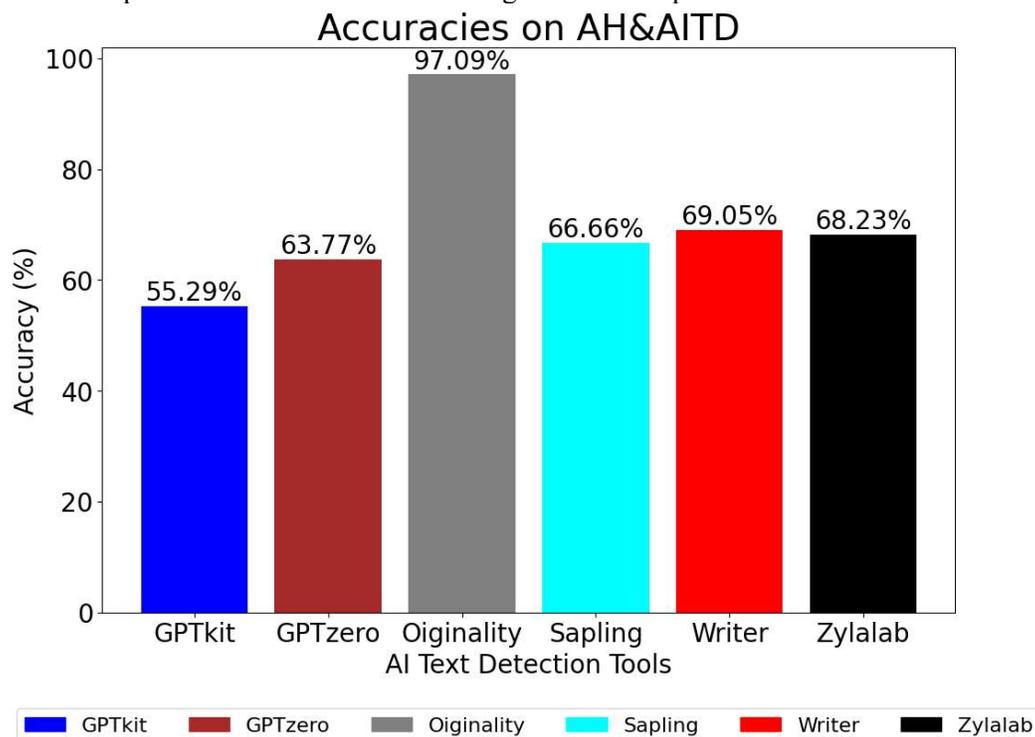

**Figure 1:** Accuracy comparison of AI text detection tools on AH&AITD

Artificial Intelligence-Generated and Human-Written Text Detection testing confusion matrices are shown in Figure 3 for simple comparison. Confusion matrices like this demonstrate visually how well certain technologies can distinguish between text generated by AI and that authored by humans. The use of blue to illustrate the matrices aids in their readability. The actual labels appear in the matrix rows, while the predicted labels are displayed in the columns. The total number of occurrences that match that criterion is in each matrix cell. These matrices allow us to compare various tools based on their ability to classify texts accurately, recall, and overall performance. This graphic is an excellent reference for consumers, researchers, and decision-makers because it visually compares the accuracy of various AI text detection technologies.

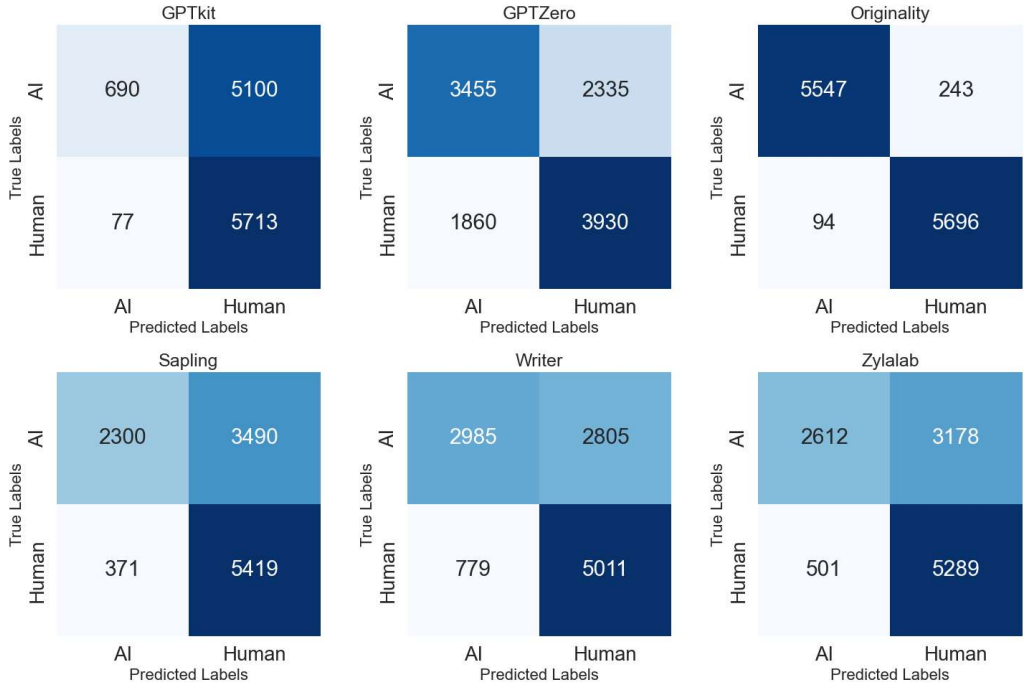

**Figure 2:** Testing confusion matrices of AI text detection tools on AH&AITD

Figure 4 displays the Testing Receiver Operating Curves (ROCs) for selecting AI text detection algorithms, visually comparing their relative strengths and weaknesses. These ROC curves, one for each tool, are essential for judging how well they can tell the difference between AI-generated and human-written material. Values for "GPTkit," "GPTZero," "Originality," "Sapling," "Writer," and "Zylalab" in terms of Area Under the Curve (AUC) are 0.55, 0.64%, 0.97%, 0.67%, 0.69%, and 0.68%, respectively. The Area under the curve (AUC) is a crucial parameter for gauging the precision and efficiency of such programs. A bigger area under the curve (AUC) suggests that the two text types can be distinguished with more accuracy. To help users, researchers, and decision-makers choose the best AI text recognition tool for their needs, Figure 4 provides a visual summary of how these tools rank regarding their discriminative strength.

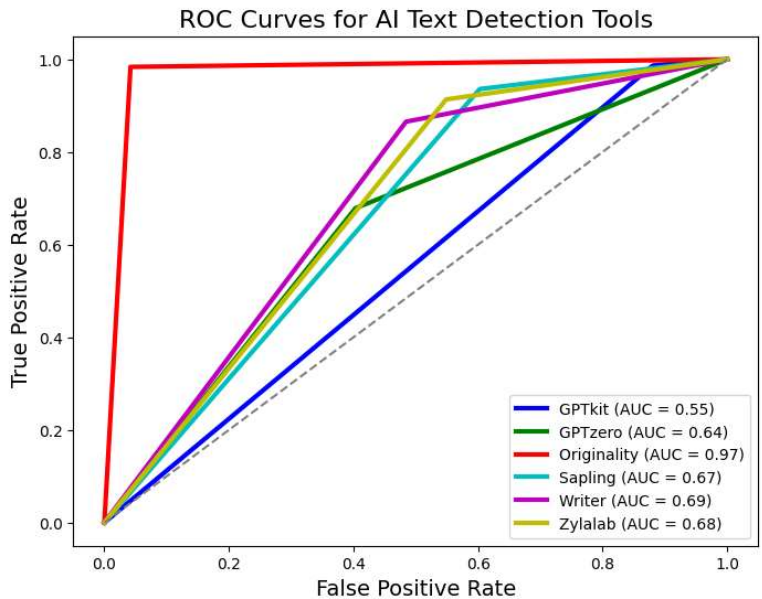

**Figure 3:** Testing Receiver Operating Curves of AI text detection tools on AH&AITD

## 5 Conclusion

We learned much about six AI-generated text identification tools by evaluating their strengths and limitations across numerous criteria using GPTkit, GPTZero, Originality, Sapling, Writer, and Zylalab. These systems have varying precision, recall, and F1 scores for distinguishing AI-generated and human-written material. Originality is impressive because it balances specificity and breadth in its effects. GPTkit excels at human-written text recognition, whereas GPTZero competes in AI-generated text precision and recall. Sapling, Writer, and Zylalab offer decent solutions. This research should focus on increasing such instruments' functions. When paraphrasing heavily, AI-generated writing must be more accurate. More work is needed to eliminate false positives because of their user impact. These tools must be more linguistically flexible and domain-specific to reach more people. Text recognition systems must keep up with AI development to detect AI-generated content in academic, creative, and social media environments. In the digital age, joint research and ongoing refinement are needed to solve new AI text detection challenges and ensure digital content accuracy.

**Availability of Data and Materials:**

Data will be provided on request. It is also publicly available.

**Conflicts of Interest:**

The authors declare no conflicts of interest to report regarding the present study.

**References**

[1] S. Baki, R. Verma, A. Mukherjee, and O. Gnawali, "Scaling and effectiveness of email masquerade attacks: Exploiting natural language generation," in *Proceedings of the 2017 ACM on Asia Conference on Computer and Communications Security*, 2017, pp. 469–482.

[2] K. Shu, S. Wang, D. Lee, and H. Liu, "Mining disinformation and fake news: Concepts, methods, and recent advancements," *Disinformation, misinformation, and fake news in social media: Emerging research challenges and opportunities*, pp. 1–19, 2020.

[3] H. Stiff and F. Johansson, "Detecting computer-generated disinformation," *International Journal of Data Science and Analytics*, vol. 13, no. 4, pp. 363–383, 2022.

[4] N. Dehouche, "Plagiarism in the age of massive Generative Pre-trained Transformers (GPT-3)," *Ethics in Science and Environmental Politics*, vol. 21, pp. 17–23, 2021.

[5] A. HLEG, "Ethics Guidelines for Trustworthy Artificial Intelligence," *High-Level Expert Group on Artificial Intelligence*, vol. 8, 2019.

[6] "Turnitin admits there are some cases of higher false positives in AI writing detection tool," K-12 Dive. Accessed: Jul. 02, 2023. [Online]. Available: https://www.k12dive.com/news/turnitin-false-positives-AI-detector/652221/

[7] C. A. de L. Salge, E. Karahanna, and J. B. Thatcher, "ALGORITHMIC PROCESSES OF SOCIAL ALERTNESS AND SOCIAL TRANSMISSION: HOW BOTS DISSEMINATE INFORMATION ON TWITTER.," *MIS Quarterly*, vol. 46, no. 1, 2022.

[8] J. Liu *et al.*, "Order-Disorder: Imitation Adversarial Attacks for Black-box Neural Ranking Models," in *Proceedings of the 2022 ACM SIGSAC Conference on Computer and Communications Security*, 2022, pp. 2025–2039.

[9] R. Schuster, C. Song, E. Tromer, and V. Shmatikov, "You autocomplete me: Poisoning vulnerabilities in neural code completion," in *30th USENIX Security Symposium (USENIX Security 21)*, 2021, pp. 1559–1575.

[10] A. Radford, J. Wu, R. Child, D. Luan, D. Amodei, and I. Sutskever, "Language models are unsupervised multitask learners," *OpenAI blog*, vol. 1, no. 8, p. 9, 2019.

[11] P. Fortuna and S. Nunes, "A survey on automatic detection of hate speech in text," *ACM Computing Surveys (CSUR)*, vol. 51, no. 4, pp. 1–30, 2018.

[12] S. Black, L. Gao, P. Wang, C. Leahy, and S. Biderman, "Gpt-neo: Large scale autoregressive language modeling with mesh-tensorflow, 2021," *If you use this software, please cite it using these metadata Search PubMed*, 2022.

[13] Y. Kashnitsky, D. Herrmannova, A. de Waard, G. Tsatsaronis, C. Fennell, and C. Labbé, "Overview of the DAGPap22 shared task on detecting automatically generated scientific papers," in *Third Workshop on Scholarly Document Processing*, 2022.

[14] W. Lu, Y. Huang, Y. Bu, and Q. Cheng, "Functional structure identification of scientific documents in computer science," *Scientometrics*, vol. 115, pp. 463–486, 2018.

[15] F. Dernoncourt and J. Y. Lee, "Pubmed 200k rct: a dataset for sequential sentence classification in medical abstracts," *arXiv preprint arXiv:1710.06071*, 2017.

[16] B. Guo et al., "How close is chatgpt to human experts? comparison corpus, evaluation, and detection," *arXiv preprint arXiv:2301.07597*, 2023.

[17] J. Wang, S. Liu, X. Xie, and Y. Li, "Evaluating AIGC Detectors on Code Content," *arXiv preprint arXiv:2304.05193*, 2023.

[18] C. A. Gao et al., "Comparing scientific abstracts generated by ChatGPT to real abstracts with detectors and blinded human reviewers," *NPJ Digital Medicine*, vol. 6, no. 1, p. 75, 2023.

[19] D. Weber-Wulff et al., "Testing of Detection Tools for AI-Generated Text," *arXiv preprint arXiv:2306.15666*, 2023.

[20] K. Krishna, Y. Song, M. Karpinska, J. Wieting, and M. Iyyer, "Paraphrasing evades detectors of ai-generated text, but retrieval is an effective defense," *arXiv preprint arXiv:2303.13408*, 2023.

[21] I. Solaiman et al., "Release strategies and the social impacts of language models," *arXiv preprint arXiv:1908.09203*, 2019.

[22] L. Kushnareva et al., "Artificial text detection via examining the topology of attention maps," *arXiv preprint arXiv:2109.04825*, 2021.

[23] R. Zellers et al., "Defending against neural fake news," *Advances in neural information processing systems*, vol. 32, 2019.

[24] E. Mitchell, Y. Lee, A. Khazatsky, C. D. Manning, and C. Finn, "Detectgpt: Zero-shot machine-generated text detection using probability curvature," *arXiv preprint arXiv:2301.11305*, 2023.

[25] S. Mitrović, D. Andreoletti, and O. Ayoub, "Chatgpt or human? detect and explain. explaining decisions of machine learning model for detecting short chatgpt-generated text," *arXiv preprint arXiv:2301.13852*, 2023.

[26] A. Pegoraro, K. Kumari, H. Fereidooni, and A.-R. Sadeghi, "To ChatGPT, or not to ChatGPT: That is the question!," *arXiv preprint arXiv:2304.01487*, 2023.

[27] D. R. Cotton, P. A. Cotton, and J. R. Shipway, "Chatting and cheating: Ensuring academic integrity in the era of ChatGPT," *Innovations in Education and Teaching International*, pp. 1–12, 2023.

[28] S. Gehrmann, H. Strobelt, and A. M. Rush, "Gltr: Statistical detection and visualization of generated text," *arXiv preprint arXiv:1906.04043*, 2019.

[29] "How can I access the ChatGPT API? | OpenAI Help Center." Accessed: Sep. 12, 2023. [Online]. Available: https://help.openai.com/en/articles/7039783-how-can-i-access-the-chatgpt-api

[30] "How can I access GPT-4? | OpenAI Help Center." Accessed: Sep. 12, 2023. [Online]. Available: https://help.openai.com/en/articles/7102672-how-can-i-access-gpt-4

[31] "OpenAI Platform." Accessed: Sep. 12, 2023. [Online]. Available: https://platform.openai.com

[32] "GPT-2 API - Developer docs, APIs, SDKs, and auth." Accessed: Sep. 12, 2023. [Online]. Available: https://apitracker.io/a/gpt-2

[33] "GPT-3 powers the next generation of apps." Accessed: Sep. 12, 2023. [Online]. Available: https://openai.com/blog/gpt-3-apps

[34] "facebook/opt-iml-30b · Hugging Face." Accessed: Sep. 12, 2023. [Online]. Available: https://huggingface.co/facebook/opt-iml-30b

[35] "FLAN-T5." Accessed: Sep. 12, 2023. [Online]. Available: https://huggingface.co/docs/transformers/model_doc/flan-t5

[36] "AI Text Detection API - API Documentation," Zyla API Hub. Accessed: Sep. 12, 2023. [Online]. Available: https://zylalabs.com/

[37] "GPTKit - AI Generated Text Detector Tool for Chat GPT," GPTKit - Highly accurate detection of GPT generated text. Accessed: Sep. 13, 2023. [Online]. Available: https://gptkit.ai

[38] "GPTZero | Frequently asked questions," GPTZero. Accessed: Sep. 13, 2023. [Online]. Available: https://gptzero.me/

[39] "Sapling," Sapling. Accessed: Sep. 13, 2023. [Online]. Available: https://blog.sapling.ai/

[40] "AI Content Checker and Plagiarism Check | GPT-4 | ChatGPT." Accessed: Sep. 15, 2023. [Online]. Available: https://originality.ai/

[41] "Introduction," Writer. Accessed: Sep. 15, 2023. [Online]. Available: https://dev.writer.com/docs

[42] "Python SDK." Writer, Aug. 14, 2023. Accessed: Sep. 15, 2023. [Online]. Available: https://github.com/writerai/writer-client-sdk-python